\documentclass{article}
\usepackage{xcolor}
\usepackage[final]{corl_2026}
\makeatletter
\renewcommand{\@noticestring}{Workshop on Robot Learning at Deployment (Learn@Deploy), CoRL 2026.}
\makeatother
\hypersetup{pdftitle={Rethinking Self-Repair in Code-as-Policy Agents: What Drives Retry Success?},
  pdfsubject={Learn@Deploy workshop, CoRL 2026},pdfauthor={Siddharth Aphale and Ayushman Singh},
  pdfkeywords={Code-as-policy agents, Self-repair, Execution feedback, Test-time evaluation}}

\usepackage[utf8]{inputenc} 
\usepackage[T1]{fontenc}    
\usepackage{url}            
\usepackage{microtype}      
\usepackage{graphicx}
\graphicspath{{./}} 
\usepackage{caption}

\usepackage{booktabs}
\usepackage{array}  
\usepackage{amsmath}

\usepackage{algorithm}
\usepackage{algpseudocode}

\usepackage{tikz}
\usetikzlibrary{positioning,arrows.meta,calc}

\usepackage{tcolorbox}

\usepackage{listings}
\tcbuselibrary{skins,breakable,listings}
\usepackage{enumitem}
\usepackage{placeins} 
\setlist[itemize]{itemsep=1pt,topsep=3pt,parsep=0pt}

\definecolor{cBgApi}{HTML}{F4F6FA}
\definecolor{cBgTask}{HTML}{EBF6EF}
\definecolor{cBgGen}{HTML}{F2F1ED}
\newtcblisting{promptpart}[2]{%
  breakable, enhanced, boxrule=0.5pt, arc=2pt, colframe=cRule, colback=#2,
  coltitle=cInk, fonttitle=\bfseries\footnotesize, title={#1},
  top=1.5pt, bottom=1.5pt, left=3pt, right=3pt, toptitle=1.5pt, bottomtitle=1.5pt,
  listing only,
  listing options={basicstyle=\scriptsize\ttfamily, breaklines=true,
    columns=fullflexible, keepspaces=true, showstringspaces=false,
    breakindent=0pt, postbreak=\mbox{\textcolor{cInkB}{$\hookrightarrow$}\,},
    aboveskip=0pt, belowskip=0pt}%
}

\definecolor{cOurs}{HTML}{7FA8CE}
\definecolor{cResample}{HTML}{9CA3AC}
\definecolor{cReflect}{HTML}{B4ACD2}
\definecolor{cMismatch}{HTML}{86B6A5}
\definecolor{cFixed}{HTML}{E1AC90}
\definecolor{cInk}{HTML}{33363C}
\definecolor{cInkB}{HTML}{656B74}
\definecolor{cRule}{HTML}{BBC0C7}
\definecolor{cBgPlan}{HTML}{EFF4FA}

\title{Rethinking Self-Repair in Code-as-Policy Agents: What Drives Retry Success?}

\author{%
  Siddharth Aphale\\
  Stanford University\\
  \texttt{saphale@stanford.edu}
  \And
  Ayushman Singh\\
  Sesame AI\\
  \texttt{ayushman@sesame.com}
}

\begin{document}
\begingroup
\makeatletter
\let\@footnotemark\H@@footnotemark
\let\@footnotetext\H@@footnotetext
\maketitle
\endgroup


\begin{abstract}
Code-as-policy agents often use execution feedback to retry failed programs. Additional sampling and task guidance can also improve success, leaving the contribution of failure-specific feedback unclear. We study this question in simulated robot manipulation. A fixed lookup selects feedback templates using failure labels and the API variant. A frozen code model generates a fresh program after each scene reset. On the same 252 initial failures, Qwen2.5-Coder-7B recovers 85.3\% within five retries, compared with 68.3\% for resampling with the same retry cap. The advantage persists under fixed total-token budgets. Feedback from other episodes and fixed task guidance yield similar recovery point estimates, with no detected benefit from matching the coarse failure label. In a separate study, a fixed guided prompt used from the first attempt achieves an observed overall success rate similar to that of template-based feedback. Template-based feedback also lowers recovery on a nut-insertion task requiring a rim grasp, consistent with an incompatible center-grasp instruction. Evaluations of robot self-repair should include resampling, mismatched feedback, and concise, task-compatible static guidance as baselines.
\end{abstract}
\keywords{Code-as-policy agents, Self-repair, Execution feedback, Test-time evaluation}

\section{Introduction}
\label{sec:intro}

Code-as-policy agents translate language instructions into executable robot programs \citep{liang2023cap}. Their failures often follow clear patterns: a program may skip the grasp, release an object after lifting, or place it away from the target. A common response is iterative self-reflection. The model reads its failed program and execution trace, writes a critique, and tries again \citep{shinn2023reflexion,madaan2023selfrefine,chen2024selfdebug}. Retry success can reflect additional sampling, concise task instructions, feedback about the current failure, or editing the previous program. We ask whether feedback after failure improves recovery over guidance available from the first attempt.

\begin{figure}[!htbp]
\centering
\begin{minipage}[t]{0.47\textwidth}
\vspace{0pt}
\makebox[\linewidth][l]{\footnotesize\textbf{(a)}}\\[3pt]
\centering
\resizebox{\linewidth}{!}{%
\begin{tikzpicture}[
  box/.style={draw=cRule, rounded corners=2pt, align=center,
              font={\fontsize{8.25}{9.5}\selectfont}, text=cInk, inner sep=5pt, text width=43mm},
  flow/.style={-{Latex[length=4pt]}, draw=cInkB, line width=0.5pt},
  note/.style={font={\fontsize{6.5}{7.5}\selectfont}, text=cInkB, inner sep=1pt},
]
\node[box] (fail) {Execution fails\\{\fontsize{6.5}{7.5}\selectfont Failure label + API variant}};
\node[box, below=3mm of fail, align=left] (feed)
  {\textbf{Select feedback}\\[3pt]
   \textcolor{cOurs}{\rule{1.5mm}{1.5mm}}\, Compiled: label lookup\\[2pt]
   \textcolor{cMismatch}{\rule{1.5mm}{1.5mm}}\, Different-label: other failure\\[2pt]
   \textcolor{cFixed}{\rule{1.5mm}{1.5mm}}\, Fixed template: task guidance\\[2pt]
   \textcolor{cResample}{\rule{1.5mm}{1.5mm}}\, Resample: no feedback};
\node[box, below=3mm of feed] (retry)
  {Reset the original scene\\Generate and execute fresh code};
\node[note, below=3mm of retry] (stop) {Success or five retries: stop};
\draw[flow] (fail) -- (feed);
\draw[flow] (feed) -- (retry);
\draw[flow] (retry) -- (stop);
\draw[flow] (retry.west) -- ++(-6mm,0) |- (feed.west);
\node[note, rotate=90, anchor=south] at ($(feed.west)!0.5!(retry.west)+(-6mm,0)$) {retry fails};
\end{tikzpicture}}
\end{minipage}\hfill
\begin{minipage}[t]{0.51\textwidth}
\vspace{0pt}
\makebox[\linewidth][l]{\footnotesize\textbf{(b)}}\\[3pt]
\includegraphics[width=\linewidth]{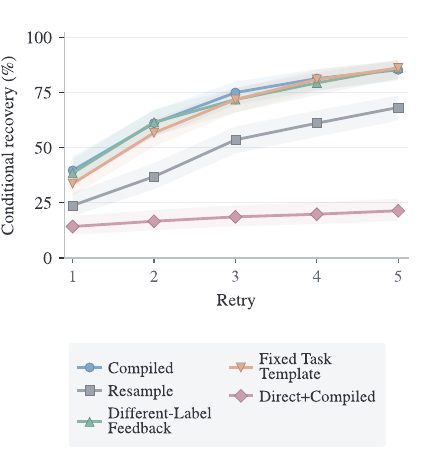}
\end{minipage}
\caption{\textbf{Template-based feedback improves recovery over resampling.} \textbf{(a)}~Reset-and-retry loop and feedback methods. \textbf{(b)}~Recovery on 252 initial failures, with similar outcomes for Compiled, Different-Label Feedback, and Fixed Task Template.}
\label{fig:main}
\end{figure}

We study this question with frozen Qwen2.5-Coder-3B/7B models in a reset-and-retry setting. Each retry starts from the original scene. Our main conditions generate a fresh program with the failed program omitted from the prompt. We use \emph{template-based feedback} to study which information improves retry success. A fixed lookup selects instructions using the failure label and API variant, such as ``MUST grasp the object before lifting.'' We call this condition \textbf{Compiled}. The lookup lets us replace matched feedback with another episode's feedback, hold the guidance fixed, or remove it while keeping the retry loop unchanged (Figure~\ref{fig:main}). Separate editing conditions expose the previous program and request a revision. All conditions restart the original scene. We do not test recovery from the physical state left by failure.

Compiled recovers 85.3\% of the 252 initial failures within five retries, compared with 68.3\% for resampling with the same retry cap. The gain persists under fixed total-token budgets. Feedback from other episodes and one fixed template per task produce similar point estimates. We detect no benefit from matching the coarse failure label to the current episode. The confidence intervals still allow differences between conditions. In a separate study, a fixed guided prompt used from the first attempt achieves an observed overall success rate similar to Compiled's, with fewer tokens per solved episode. That comparison changes initial guidance and the generation instruction together. Their individual contributions remain unresolved.

We compare feedback content and program-generation strategies under matched retry budgets to test how additional sampling, reusable task guidance, and failure-label matching affect recovery. Evaluations of robot self-repair should include static guidance as a baseline for adaptive feedback.

\section{Related Work}
\label{sec:related}

\textbf{Language models for robot control.} Code as Policies and ProgPrompt generate executable programs from grounded robot skills \citep{liang2023cap,singh2023progprompt}. SayCan selects actions using language-model and affordance scores \citep{ahn2022saycan}. CaP-X evaluates code-generating agents across robot APIs and feedback modalities, including multi-turn interaction without scene resets \citep{capx}. Robot recovery systems often ask a model to interpret a failure. REFLECT summarizes robot experience to support failure explanation and correction \citep{liu2023reflect}. DoReMi compares a plan with its execution \citep{guo2023doremi}. TCoT adds trajectory-level reasoning to a vision-language-action model \citep{li2026tcot}. Our reset-based comparisons examine feedback content and program-generation strategies on paired episodes under matched retry budgets.

\textbf{Evaluation of iterative self-repair.} Reflexion and Self-Refine use model-written critiques to improve later outputs \citep{shinn2023reflexion,madaan2023selfrefine}. Self-Debugging conditions code generation on the previous program and its execution results \citep{chen2024selfdebug}. On code benchmarks, \citet{olausson2024selfrepair} found that self-repair often does not outperform independent sampling at matched compute. Self-correction of reasoning is also unreliable without external feedback, whereas critique grounded in external evidence can help \citep{huang2024selfcorrect,gou2024critic}. We extend matched-sampling comparisons to embodied code execution and test whether guidance must match the current failure or follow a failure at all.

\textbf{Execution feedback and interactive evaluation.} CodeRL, RLEF, and MURPHY use compiler or test feedback to update model parameters \citep{le2022coderl,gehring2024rlef,murphy}. RLEF also tests random execution feedback from unrelated programming problems at inference time. This comparison tests the value of feedback specific to the current problem, although its training, tasks, and feedback construction differ from ours. Inference-time agents such as ReAct, Voyager, and LATS use feedback during search or interaction \citep{yao2023react,wang2023voyager,zhou2024lats}. InterCode, AppWorld, and AgentBoard evaluate execution outcomes and intermediate progress \citep{yang2023intercode,trivedi2024appworld,ma2024agentboard}. With model weights fixed, we ask which information in a robot's retry prompt accounts for recovery.

\section{Methods}
\label{sec:method}

\subsection{Experimental Setup}

We evaluate frozen code models in robosuite simulation with a Panda arm \citep{zhu2020robosuite,todorov2012mujoco}. Models receive state information and generate Python programs using a documented robot API, without images. The primary evaluation uses Qwen2.5-Coder-7B-Instruct \citep{hui2024qwencoder} with the primitive API on cube lifting and stacking.

An \emph{episode} comprises an initial program execution for a task and scene, followed by any retries under one method. The initial execution is \emph{attempt~0}. After an initial failure, the model is allowed up to five \emph{retries}. Before each retry, the simulator resets to the original scene. Each retry produces a complete program, either generated from scratch or revised from the previous program.

We use 252 initial failures for evaluation and 40 for development. The sets contain distinct initial programs, identified by their hashes. The evaluation covers 159 combinations of task, API variant, and scene, including 24 also used in development. All methods are evaluated on the same stored initial failures. The end-to-end studies pair methods by scene seed.

\emph{Conditional recovery@$k$} is the fraction of attempt-0 failures solved within $k$ retries. This metric measures sequential retries and differs from pass@$k$ estimated from independent samples. For end-to-end studies, \emph{initial success} is success on attempt~0 and \emph{overall success@6} is success in any of the six allowed executions. An attempt is \emph{valid} if its program compiles, uses only documented API functions, and finishes before timeout.

The generalization study includes Qwen2.5-Coder-3B/7B and two additional pick-and-place tasks. We combine results across seven settings that meet the predefined eligibility criteria. Separate studies examine API variants with fewer high-level helper functions, Qwen2.5-Coder-32B, and DeepSeek-Coder-6.7B \citep{guo2024deepseekcoder}. Further details of the software, tasks, evaluation sets, and metrics are provided in Appendix~\ref{sec:app-setup}.

\subsection{Template-Based Feedback}

After an unsuccessful execution, a rule-based labeler assigns one primary label from a 14-label taxonomy. \textbf{Compiled} maps this label and the API variant to a fixed \emph{feedback template} containing a diagnosis and short MUST/NEVER instructions (Table~\ref{tab:templates}). This lookup requires no additional model call. The instructions guide code generation without imposing execution-time checks.

Compiled accumulates feedback from the initial failure and subsequent failed retries, retaining the labels and attempt indices. The \emph{feedback block} presents the newest feedback first, omits duplicate instruction lines, and respects the feedback-length limit. The generation prompt contains the task, API documentation, and accumulated feedback, with the previous program omitted. An example from the earlier study shows an initial failure followed by a successful fresh retry (Figure~\ref{fig:filmstrip}a). The prompts are provided in Appendix~\ref{sec:app-prompts}, and the reset-and-retry algorithm is given in Algorithm~\ref{alg:loop}.

\begin{table}[!t]
\centering
\caption{\textbf{Failure labels select fixed retry guidance.} Example instruction excerpts for the primitive API. Full templates are provided in Appendix~\ref{app:full-library}.}
\label{tab:templates}
\small
\setlength{\tabcolsep}{4pt}
\begin{tabular}{@{}>{\raggedright\arraybackslash}p{0.30\linewidth} >{\raggedright\arraybackslash}p{\dimexpr0.70\linewidth-8pt\relax}@{}}
\toprule
Detected failure & Guidance added to the retry prompt\\
\midrule
Grasp did not secure the object & MUST: grasp at the object's CENTRE: \ldots, with the gripper empty\\
\addlinespace
Object grasped but not lifted & MUST: \ldots{} call \texttt{lift(name, height=0.1)} after the grasp\\
\addlinespace
Unsupported function call & MUST: use ONLY the documented API functions \ldots\newline NEVER: import statements\\
\bottomrule
\end{tabular}
\end{table}

\subsection{Comparison Methods}

We compare methods within each study using the same generator, task, API, decoding settings, and retry limit. \textbf{Resample} repeats the base retry prompt without feedback. \textbf{Other-Episode Feedback} replaces the current episode's feedback with feedback from other failures, matched by task, API variant, and initial task-progress bin. The primary matching control, \textbf{Different-Label Feedback}, selects individual feedback records from other episodes of the same task, preferring the same initial task-progress bin and requiring a different failure label whenever available. Both add feedback as the evaluated episode accumulates failures and use the same feedback-length limit. These controls test the value of matching the coarse failure label while retaining useful task information. The matching and fallback rules are specified in Appendix~\ref{app:control-definitions}.

We select a fixed template for each task from the development data, choosing the most frequent failure label with task-compatible guidance. \textbf{Fixed Task Template} reuses this template across retries after the initial failure. \textbf{Static-from-Start} uses the complete Fixed Task Template retry prompt from attempt~0, changing both the initial generation instruction and the presence of guidance.

\textbf{Direct+Compiled} shows the failed program and requests an edit, jointly changing code exposure and the generation instruction. \textbf{Reflection} uses an extra model call to critique the failed program and execution log. Its generation call receives the accumulated critiques and omits the previous code. Reflection was tested only in the earlier core evaluation with the original generation instruction.

The primary evaluation uses a condition-neutral generation instruction for regeneration methods. The original instruction referred to a repair plan even when feedback was absent. Both prompt versions are provided in Appendix~\ref{app:prompt-format}.

Eligibility criteria, multiplicity corrections, decoding settings, and token accounting are specified in Appendix~\ref{sec:app-setup}.

\begin{figure}[!t]
\centering
\makebox[\linewidth][l]{\footnotesize\textbf{(a)}}\\[2pt]
\includegraphics[width=\linewidth]{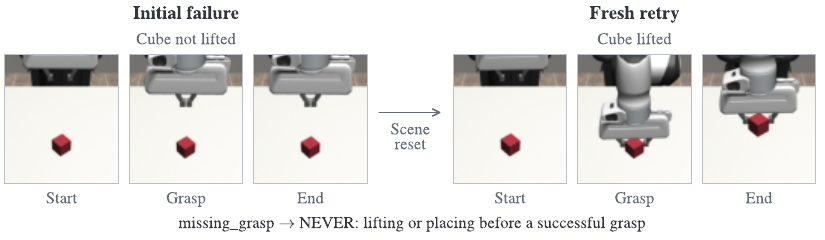}\\[4pt]
\includegraphics[width=\linewidth]{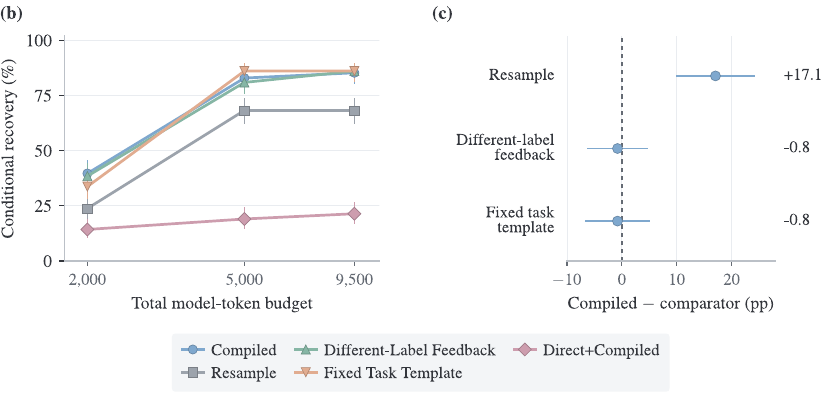}
\caption{\textbf{Guidance improves recovery under fixed token budgets.} \textbf{(a)}~An initial failure followed by a successful retry after reset. \textbf{(b)}~Recovery under total-token budgets. \textbf{(c)}~Paired recovery@5 comparisons show an advantage over resampling, with no detected advantage over different-label or fixed-template guidance.}
\label{fig:filmstrip}
\label{fig:budget}
\label{fig:attribution}
\end{figure}

\section{Results}
\label{sec:results}

\subsection{Recovery under Matched Budgets}
\label{sec:mainresults}

In the primary evaluation, Compiled recovers 85.3\% of the 252 held-out failures within five retries. Resample, with the same retry cap, recovers 68.3\%, a difference of $+17.1$\,pp (95\% CI $[+9.9,+24.2]$).\footnote{pp denotes percentage points. Bands and error bars show 95\% CIs: Wilson for recovery proportions and paired bootstrap for differences.} Compiled recovers more failures than Resample from the first retry (Figure~\ref{fig:main}b) and uses fewer total model tokens, including prompts and completions.

Under fixed token budgets, Compiled improves recovery over Resample by $+15.9$\,pp at 2{,}000 tokens and $+14.7$\,pp at 5{,}000 (Figure~\ref{fig:budget}b). Fixed Task Template and Different-Label Feedback yield similar observed recovery rates to Compiled, while Direct+Compiled has lower recovery. Compiled also achieves higher overall success than Resample in the earlier 700-scene end-to-end study with the original generation instruction (Appendix~\ref{sec:app-results}).

\subsection{Feedback Matching and Content}
\label{sec:mechanism}

Different-Label Feedback and Fixed Task Template each recover 86.1\% in the primary evaluation, compared with 85.3\% for Compiled. We detect no advantage from matching the coarse label to the current failure, although the paired confidence intervals do not establish equivalence (Table~\ref{tab:contrasts}; Figure~\ref{fig:attribution}c). In earlier studies, we also detected no advantage for Compiled over Other-Episode Feedback, Permuted Labels, or Initial-Label Only (Appendix~\ref{sec:app-results}).

In the primary evaluation, the exploratory Repeated API Rules control copies rules already in the prompt and does not reproduce Compiled's improvement. Earlier comparisons using the original generation instruction likewise show lower recovery with content-free feedback and All-Template Union than with concise feedback templates (Appendix~\ref{sec:app-results}).

\begin{table}[!htbp]
\centering
\caption{\textbf{Primary paired contrasts} on 252 initial failures. Each entry is Compiled minus the comparator in recovery@5, with a 95\% paired-bootstrap CI.}
\label{tab:contrasts}
\small
\setlength{\tabcolsep}{3pt}
\begin{tabular}{@{}>{\raggedright\arraybackslash}p{0.28\columnwidth} >{\raggedright\arraybackslash}p{0.36\columnwidth} l@{}}
\toprule
Comparator & Question & $\Delta$ (pp; 95\% CI)\\
\midrule
Resample & benefit beyond another sample? & $+17.1$ $[+9.9,+24.2]$\\
Different-Label Feedback & must the label match? & $-0.8$ $[-6.3,+4.8]$\\
Fixed Task Template & dynamic versus fixed guidance? & $-0.8$ $[-6.7,+5.2]$\\
Direct+Compiled & does editing improve recovery? & $+63.9$ $[+56.7,+70.6]$\\
\bottomrule
\end{tabular}
\end{table}

\subsection{Guidance from the First Attempt}

On 700 new scenes, Static-from-Start increases initial success from 19.3\% with the standard prompt to 36.9\%. It uses the complete Fixed Task Template retry prompt from the first attempt, changing both task guidance and the generation instruction. Their individual contributions remain unresolved.

\begin{figure}[!htbp]
\centering
\includegraphics[width=\linewidth]{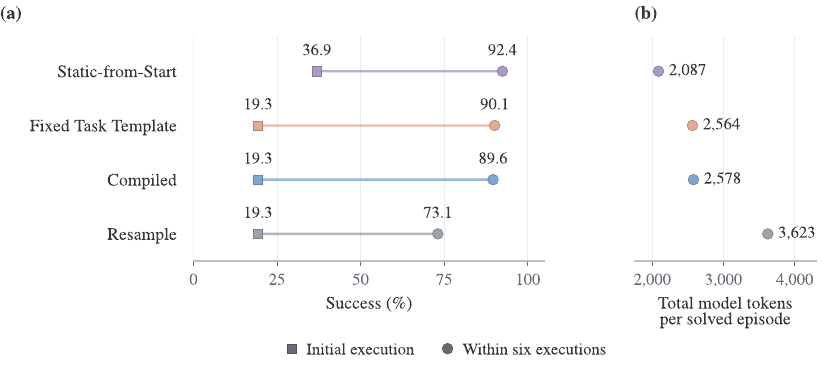}
\caption{\textbf{A fixed guided prompt improves initial success.} \textbf{(a)}~Initial and overall success on 700 fresh scenes. \textbf{(b)}~Total model tokens per solved episode. Static-from-Start changes both the initial instruction and guidance.}
\label{fig:static-start}
\end{figure}

By the sixth execution, Static-from-Start reaches 92.4\% overall success, compared with 89.6\% for Compiled and 90.1\% for Fixed Task Template (Figure~\ref{fig:static-start}). Its advantage over Compiled is not statistically significant after multiplicity correction ($+2.9$\,pp; Holm-adjusted $p{=}0.13$). It also uses the fewest total model tokens per solved episode. Complete results and prompt differences are reported in Appendix~\ref{app:static-start}.

The matching controls and Static-from-Start results suggest that reusable task guidance accounts for the gain. These results do not establish that the model uses the current execution trace.

\subsection{Editing and Reflection}

Direct+Compiled achieves 21.4\% recovery in the primary evaluation. Earlier analyses show smaller program changes with editing than with fresh generation. Because Direct+Compiled changes code exposure and the generation instruction together, the cause of its lower recovery remains unresolved.

In the earlier core evaluation, Reflection achieves 56.0\% recovery, below Resample, while using more completion tokens and wall-clock time. These findings concern the tested editing and reflection implementations. Further editing and critique results are summarized in Appendix~\ref{app:editing-specificity}.

\subsection{Generalization and Failure Cases}
\label{sec:boundaries}

\begin{figure}[!t]
\centering
\includegraphics[width=\linewidth]{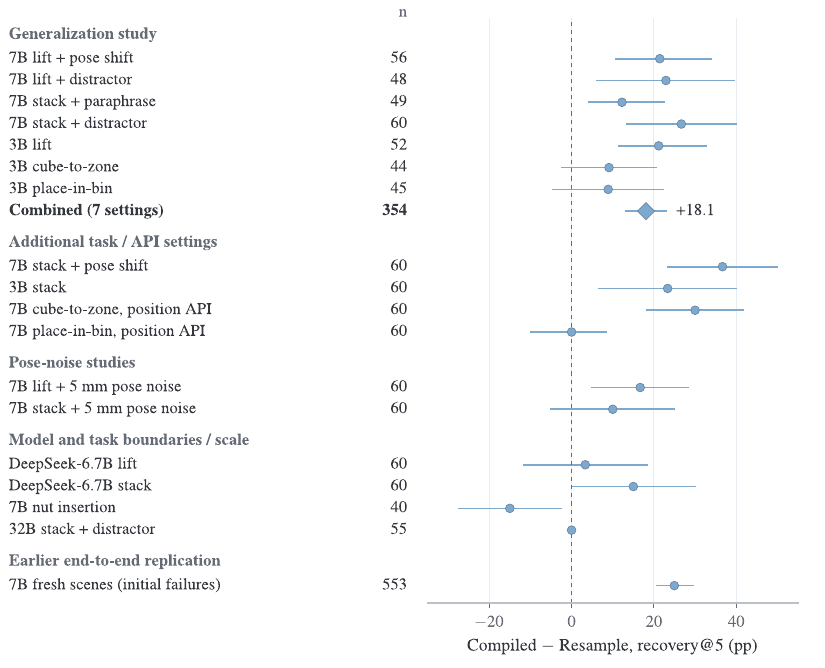}
\caption{\textbf{Guidance benefits vary across tasks and models.} Compiled--Resample recovery differences across evaluated settings. The diamond summarizes the seven eligible generalization settings.}
\label{fig:forest}
\end{figure}

Across seven Qwen2.5-Coder-3B/7B settings covering four tasks and 354 initial failures, Compiled improves recovery@5 over Resample by $+18.1$\,pp (CI $[+13.3,+22.9]$). Its difference from Other-Episode Feedback is smaller, at $+2.3$\,pp (CI $[-1.1,+5.6]$), with no detected advantage over this same-task feedback control (Figure~\ref{fig:forest}; Appendix~\ref{sec:app-generalization}).

The recovery advantage persists in most additional evaluations of API variants, pose noise, and generation seeds. At temperature 0.2 on lifting with pose shift, however, Compiled's recovery is 19.6\,pp lower than Resample's.

With Qwen2.5-Coder-32B, Compiled and Resample solve the same 52 of 55 initial stack failures (94.5\% recovery@5). This high recovery rate leaves limited room for further improvement.

On nut insertion, the required rim grasp conflicts with the center-grasp instruction in the pick-and-place template for \texttt{failed\_grasp}. Recovery@5 is 82.5\% with Compiled and 97.5\% with Resample, a difference of $-15.0$\,pp (CI $[-27.5,-2.5]$). The decrease is consistent with the incompatible grasp instruction, while attempt validity remains similar. Guidance that helps within one task family can therefore be harmful in another.

\section{Discussion}
\label{sec:discussion}

Our results suggest that reusable task guidance explains the recovery advantage of template-based feedback over resampling in the tested tasks. Fixed guidance and feedback from other failures yield similar recovery point estimates to Compiled, with no detected benefit from matching the coarse failure label. Static-from-Start also achieves similar observed overall success to Compiled while using fewer tokens per solved episode. Evaluations should test fixed guidance from the first attempt alongside feedback selected after failure. The Static-from-Start comparison changes both guidance and the generation instruction, so their individual contributions remain unresolved.

Evaluations of self-repair should therefore distinguish the contributions of additional sampling, reusable guidance, and failure-specific information. Resampling provides a baseline for additional sampling, with retry and token budgets matched across methods. Static guidance tests whether instructions are useful before a failure occurs, while mismatched feedback tests the value of matching the coarse failure label to the current episode. These comparisons should check for instructions that conflict with a task's requirements, such as center-grasp guidance on a task requiring a rim grasp.

\section{Conclusion}
\label{sec:conclusion}

Evaluations of robot self-repair should include resampling, mismatched feedback, static guidance from the first attempt, and previous-program editing, with retry and token budgets matched across methods.

Our findings are limited to simulated manipulation with privileged state and coarse feedback. Some comparisons combine instruction and information changes. Future work should isolate these effects and examine larger models on harder tasks or under tighter budgets. Extending evaluation to visual feedback and physical recovery, including post-failure states and action costs, would clarify when adaptive feedback improves on static guidance.

\bibliography{main}

\clearpage
\appendix
\setcounter{table}{0}
\setcounter{figure}{0}
\renewcommand{\thetable}{A\arabic{table}}
\renewcommand{\thefigure}{A\arabic{figure}}
\renewcommand{\theHtable}{A\arabic{table}}
\renewcommand{\theHfigure}{A\arabic{figure}}

\section{Experimental Details}
\label{sec:app-setup}

\subsection{Tasks, APIs, and evaluation sets}
\label{app:environment}

We evaluate text-only code generation with privileged state in robosuite~1.5.2 and MuJoCo~3.3.5 using a simulated Panda arm. Python programs run through a documented position-control API in isolated subprocesses with a 60\,s timeout. The primitive API includes \texttt{lift} and \texttt{place\_on}; the position-only API omits them. The tasks are cube lifting, stacking, cube-to-zone, and place-in-bin, with additional evaluations on nut insertion and door opening.

The evaluation sets, retry instructions, and outcome measures are summarized in Table~\ref{tab:app-studies}. The primary dataset contains 360 Qwen2.5-Coder-7B programs (18.9\% initial success). Its 292 failures unrelated to infrastructure errors are divided into 40 development and 252 evaluation failures by initial-program hash.

Within each study, eligibility requires initial success in the predefined range $[10\%,70\%]$. The combined generalization estimate weights the 354 initial failures from seven eligible settings equally. Settings outside this range and separately designed studies are reported separately.

\begin{table}[!htbp]
\centering
\caption{\textbf{Evaluation sets and endpoints.} Comparisons pair conditions within each study. Original and neutral instructions are reproduced in Appendix~\ref{app:prompt-format}.}
\label{tab:app-studies}
\footnotesize
\setlength{\tabcolsep}{4pt}
\begin{tabular}{@{}>{\raggedright\arraybackslash}p{0.20\textwidth}>{\raggedright\arraybackslash}p{0.36\textwidth}>{\raggedright\arraybackslash}p{0.15\textwidth}>{\raggedright\arraybackslash}p{0.21\textwidth}@{}}
\toprule
Study & Model and evaluation set & Retry instruction & Main endpoint\\
\midrule
Primary & Qwen2.5-Coder-7B; 252 held-out failures, lifting and stacking & Neutral & Recovery within five retries\\
\addlinespace
Generalization & Qwen2.5-Coder-3B/7B; 354 failures across seven eligible settings, four tasks & Original & Recovery within five retries\\
\addlinespace
Earlier end-to-end & Qwen2.5-Coder-7B; 700 fresh scenes, lifting and stacking & Original & Overall success within six executions\\
\addlinespace
Static-from-Start & Qwen2.5-Coder-7B; 700 separate fresh scenes, lifting and stacking & Neutral & Overall success within six executions\\
\bottomrule
\end{tabular}
\end{table}

\subsection{Decoding, metrics, and statistical analysis}
\label{app:inference}
\label{app:retry-protocol}

After an initial failure, each condition receives at most five retries from the original scene. Within a study, conditions share the model, task, API, decoding settings, and retry cap. Unless stated otherwise, generation uses temperature 0.7, top-$p$ 0.95, a 1{,}024-token completion limit, and a 768-whitespace-token feedback cap. Reflection calls allow 256 completion tokens.

Conditional recovery@$k$ is the fraction of initial failures solved within $k$ sequential retries. Task success is checked by an independent final-state geometric predicate. Valid programs compile, call only documented functions, and finish before timeout. An execution counts as a safety violation if it leaves an object off the table.

Confidence intervals for paired differences use 10{,}000 bootstrap replicates. Exact McNemar tests assess significance at the final retry, with Holm correction within each predefined comparison group. Intervals are 95\% unless an earlier secondary result is marked 90\%. The primary analysis treats each stored initial failure as one observation. An exploratory bootstrap over the 159 task--API--scene clusters accounts for dependence within scene configurations and retains the $+17.1$\,pp Compiled--Resample difference (95\% CI $[+10.1,+24.2]$).

Fixed-token analyses replay logged calls in order. Each call's prompt and maximum completion allowance must fit within the remaining budget. Token use includes prompts, completions, and reflection calls. The budgets are 2{,}000, 5{,}000, and 9{,}500 tokens. No call fits within the original 1{,}000-token budget. In end-to-end studies, tokens per solved episode is total token use across all scenes, including attempt~0 and unsolved scenes, divided by the number of solved episodes.

\subsection{Reproducibility notes}
\label{app:pairing}

Development and evaluation contain distinct initial programs but share 24 of the 159 evaluation scene configurations. The earliest core comparisons pair stored initial failures, though a placement-sampler seeding limitation allowed retry layouts to differ. Later studies seed every reset, and the subsequent evaluation with the original generation instruction uses rebuilt software. Each comparison uses a single software build. Seeding does not guarantee bitwise deterministic simulation.

The generalization question was revised after the core results, with settings and analysis fixed before generation for that study. The Static-from-Start study was also specified before generation. Fixed Task Template was selected on development data, so its first held-out comparison is exploratory. Later studies reused this fixed text. Direct+Compiled was selected in a separate 116-failure pilot.

Failure labels were audited on scripted faults, without human audits of natural failures. Oracle-hint performance and template-authoring cost were not measured.

\section{Feedback and Prompt Construction}
\label{sec:app-prompts}

\subsection{Failure labels and template selection}
\label{app:compiler}

The labeler checks execution errors, safety events, absence of API calls, and task subgoals in a fixed priority order to assign one of 14 failure labels. The lookup $\textsc{Compile}(\mathit{label},\mathit{variant})$ returns a diagnosis and MUST/NEVER instructions. Three templates vary across APIs, yielding 17 label-specific texts. All templates, API substitutions, and the generic fallback are reproduced in Appendix~\ref{app:full-library}.

Compiled adds one feedback record after each failure. Records are ordered newest first, with duplicate instruction lines removed and the shared feedback-length limit applied. Attempt indices remain in the headers, so repeated labels can appear under different attempt headers.

\subsection{Reset-and-retry algorithm}

Regeneration with Compiled, mismatched feedback, and fixed guidance is described in Algorithm~\ref{alg:loop}, with selection rules in Appendix~\ref{app:control-definitions}. Each failure record includes its label and the information used to match source episodes.

\begin{algorithm}[!htbp]
\caption{Reset-and-retry regeneration with condition-specific feedback}
\label{alg:loop}
\footnotesize
\begin{algorithmic}[1]
\Require original scene $s_0$, initial failure record $r_0$, API variant $t$, condition $a$, retry cap $K$, feedback cap $B$
\State $\mathcal{R}\gets\langle r_0\rangle$ \Comment{evaluated episode's failures}
\For{$i=1$ to $K$}
  \State $\mathcal{H}\gets\textsc{SelectFeedback}(a,\mathcal{R},t)$ \Comment{empty for Resample}
  \State $f\gets\textsc{RenderFeedback}(\mathcal{H},B)$ \Comment{newest first; deduplicate instructions}
  \State $x\gets\textsc{RenderPrompt}(\textrm{task},\textrm{API}(t),f)$
  \State $\textsc{Reset}(s_0)$; $p\gets\textsc{Generate}(x)$ \Comment{previous program hidden}
  \State $(\mathit{success},r)\gets\textsc{Execute}(p)$
  \If{$\mathit{success}$} \State \Return \textsc{Success} \EndIf
  \State $\mathcal{R}\gets\mathcal{R}\mathbin{\Vert}\langle r\rangle$
\EndFor
\State \Return \textsc{Fail}
\end{algorithmic}
\end{algorithm}

\subsection{Control definitions}
\label{app:conditions}
\label{app:control-definitions}

Regeneration prompts omit the previous program (Table~\ref{tab:app-conditions}). Direct+Compiled and Reflection Repair include it and request a revision, changing both code exposure and the generation instruction.

\begin{table}[!htbp]
\centering
\caption{\textbf{Prompt inputs for each feedback and editing condition.} Source-episode selection and prompt construction are specified below.}
\label{tab:app-conditions}
\footnotesize
\setlength{\tabcolsep}{4pt}
\begin{tabular}{@{}>{\raggedright\arraybackslash}p{0.29\textwidth}>{\raggedright\arraybackslash}p{0.67\textwidth}@{}}
\toprule
Condition & Feedback supplied to program generation\\
\midrule
Resample / Compiled & No feedback / accumulated templates for the observed labels.\\
Other-Episode Feedback & Accumulated feedback from a matched source episode.\\
Different-Label Feedback & Feedback from another failure, prioritizing a different label.\\
Permuted Labels & A fixed permutation replaces every observed label with a different label.\\
Cross-Task Feedback & A fixed assignment of source episodes from a different task.\\
Fixed Task Template & One fixed development-selected template per task, reused after failure.\\
Static-from-Start & The complete Fixed Task Template retry prompt from attempt~0 onward.\\
Initial-Label Only & The initial failure's template, unchanged across retries.\\
All-Template Union & Goal-compatible templates ordered by development frequency.\\
Static Prior & One fixed task-independent text per model, without accumulation.\\
Placebo / Repeated API Rules & Length-matched generic instructions / existing API execution rules.\\
Direct+Compiled & Previous program and accumulated feedback templates.\\
Reflection & Accumulated model-written critiques; previous program hidden.\\
Reflection Repair & Previous program, latest critique, and truncated execution log.\\
\bottomrule
\end{tabular}
\end{table}

Other-Episode Feedback selects feedback from another episode matched by task, API variant, and initial task-progress bin. Progress is the fraction of ordered subgoals completed before the first unmet subgoal, grouped into four equal-width bins over $[0,1]$. Source histories are limited to the evaluated episode's failure count.

Different-Label Feedback selects one source record per observed label from the same task, preferring a different label in the same progress bin. If none is available, the progress-bin match is relaxed before allowing the same label; the task match is always retained. In the primary evaluation, all 627 selected records had a different failure label from the corresponding observed failure. Both controls use Compiled's feedback ordering, deduplication, and length limit.

Fixed Task Template uses the most frequent failure label with task-compatible guidance for each task, selected from the fixed 40-failure development set. Static-from-Start also reuses the retry instruction and historical failure header (Appendix~\ref{app:static-start}).

Following Reflexion \citep{shinn2023reflexion}, Reflection uses a separate critic call that receives the previous program and truncated log. Generation receives only the accumulated critique.

\subsection{Prompt construction}
\label{app:prompt-format}

A generation prompt concatenates four blocks:
\begin{center}
\small
\fbox{API documentation} $\rightarrow$ \fbox{task and objects} $\rightarrow$ \fbox{condition-specific feedback} $\rightarrow$ \fbox{generation instruction}.
\end{center}
The API and task blocks below are the primitive-tier stacking example; only the feedback block varies across regeneration conditions.

\begin{promptpart}{API documentation and execution rules (identical in every condition and attempt at a given tier)}{cBgApi}
# API
Code-as-policy API (privileged state, position control; all positions are (x, y, z)
tuples in metres in the world frame):
  get_object_pose(name, return_bbox_extent=False) -> (x, y, z)
      # object centre position; with return_bbox_extent=True returns
      # ((x, y, z), (ex, ey, ez)) where e* are half-extents of the object
  get_gripper_pose() -> (x, y, z)      # current gripper fingertip position
  goto_pose(position, quat=None, z_approach=0.10)
      # move the gripper to `position`, approaching from z_approach metres above;
      # orientation is fixed top-down (quat is accepted but ignored)
  open_gripper()                       # open the fingers (release)
  close_gripper()                      # close the fingers (grasp whatever is between them)
  check_grasped(name) -> bool          # True if `name` is currently held
  execute_waypoints(waypoints)         # goto_pose each position in the list, in order
  lift(name, height=0.1)               # raise the HELD object by height metres
  place_on(source, target)             # move the HELD source above target and release it

Execution rules:
  - These functions are ALREADY DEFINED in the execution environment. Do not import
    anything and do not define or redefine any of these functions yourself.
  - Top-level code runs directly; if you wrap steps in a helper function you must call it.
  - To grasp an object: goto_pose to its centre position, then close_gripper().
  - Success is checked on the FINAL state of the scene: a lifted object must still be
    held at the end (do not call open_gripper() after lifting unless the task says to
    release); to place an object, move it above the target and open_gripper().
  - place_on(source, target) positions the held object on the target and
    releases it automatically; do not call open_gripper() after place_on.
\end{promptpart}
\begin{promptpart}{Task and objects}{cBgTask}
# Task
Stack the red cube on top of the green cube.

# Objects
red_cube, green_cube
\end{promptpart}
For the primitive API tier, releasing a lifted object produces the following exact block:

\begin{promptpart}{Condition-specific feedback example}{cBgPlan}
# Repair plan
attempt 0 failed: released_after_lift - The object was lifted but then released, so it was no longer held at the end.
  MUST: do NOT open the gripper after lifting; keep holding (or place on the target first): use place_on(source, target) as the final action (it releases)
  NEVER: open_gripper immediately after lifting
\end{promptpart}

The primary evaluation used the following instruction for regeneration conditions:

\begin{promptpart}{Condition-neutral generation instruction}{cBgGen}
# Write a COMPLETE fresh Python program using ONLY the API above, following the same execution rules (no imports, no redefining API functions, call any helper you define). Follow every instruction present in this prompt. Output only code.
\end{promptpart}

The position-only tier removes \texttt{lift} and \texttt{place\_on} from the API block. Earlier runs used a suffix that referred to a ``repair plan,'' \texttt{must\_change} items, and a verification checklist even when those fields were absent. We therefore repeated the core comparisons with the neutral instruction above; the main paper reports that evaluation. The original generation instruction was:
\begin{promptpart}{Historical generation instruction}{cBgGen}
# Using the repair plan above (compiled from a previous failed attempt), write a COMPLETE fresh Python program using ONLY the API above, following the same execution rules (no imports, no redefining API functions, call any helper you define). Satisfy every must_change item and the verification checklist. Output only code.
\end{promptpart}
In the primary evaluation, Direct+Compiled uses the following editing instruction:
\begin{promptpart}{Neutral editing instruction}{cBgGen}
# Revise the previous program above into a COMPLETE corrected Python program using ONLY the API above, following the same execution rules (no imports, no redefining API functions, call any helper you define). Follow every instruction present in this prompt. Output only code.
\end{promptpart}
The previous program is included only in the editing prompt. Reflection uses a separate call with the instruction ``In at most 6 sentences, reflect on why the program failed the task and what to do differently next time. Output only the reflection.'' Serialized prompts and model revisions are recorded in the experiment artifacts.

\section{Additional Analyses}
\label{sec:app-results}

\subsection{Feedback content and failure matching}
\label{app:content}
\label{app:editing-specificity}

\begin{figure}[!htbp]
\centering
\begin{minipage}[t]{0.85\textwidth}
\centering
\makebox[\linewidth][l]{\footnotesize\textbf{(a)}}\\[3pt]
\includegraphics[width=\linewidth]{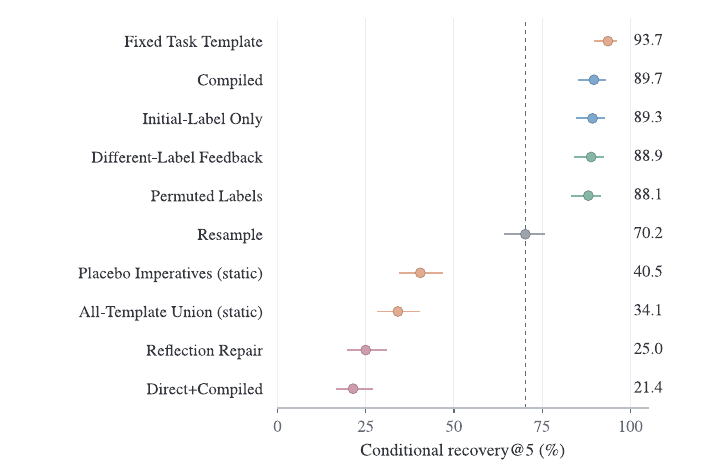}
\end{minipage}\par\medskip
\begin{minipage}[t]{0.85\textwidth}
\centering
\makebox[\linewidth][l]{\footnotesize\textbf{(b)}}\\[3pt]
\includegraphics[width=\linewidth]{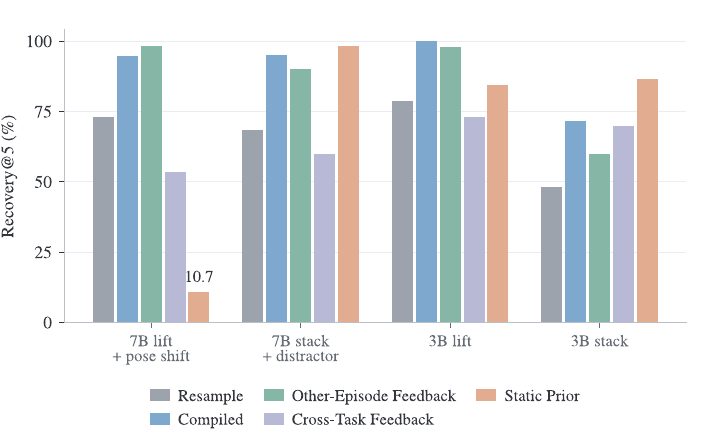}
\end{minipage}
\caption{\textbf{(a)}~Recovery with feedback and editing conditions in earlier evaluations. \textbf{(b)}~A task-independent prompt improves recovery on stacking but reduces it on lifting.}
\label{fig:mechanism}
\end{figure}

We detect no advantage from matching the coarse failure label in the primary evaluation. The paired recovery difference between Compiled and Different-Label Feedback is $-0.8$\,pp (95\% CI $[-6.3,+4.8]$). Earlier comparisons using the original generation instruction also show little benefit from updating the initial label or preserving the label mapping (Table~\ref{tab:app-attribution}). These intervals do not establish equivalence.

In these earlier comparisons, recovery is lower with content-free instructions and All-Template Union than with concise, task-compatible feedback (Figure~\ref{fig:mechanism}). The exploratory Repeated API Rules control also yields lower recovery than Resample in the primary evaluation (61.1\% versus 68.3\%).

\begin{table}[!htbp]
\centering
\caption{\textbf{Additional controls test feedback content and label matching.} Comparisons using the original generation instruction on 252 held-out failures. $p$-values are Holm-adjusted within predefined groups of secondary comparisons.}
\label{tab:app-attribution}
\footnotesize
\setlength{\tabcolsep}{5pt}
\begin{tabular}{@{}lrrr@{}}
\toprule
Contrast & $\Delta$ (pp) & 95\% CI & $p$\\
\midrule
Reflection Repair $-$ Resample & $-45.2$ & $[-53.2,-36.9]$ & $2.9\times10^{-21}$\\
All-Template Union $-$ Resample & $-36.1$ & $[-44.0,-27.8]$ & $3.6\times10^{-14}$\\
Compiled $-$ Initial-Label Only & $+0.4$ & $[-4.0,+4.8]$ & $1.0$\\
Compiled $-$ Permuted Labels & $+1.6$ & $[-3.6,+6.7]$ & $1.0$\\
Placebo $-$ Resample & $-29.8$ & $[-37.7,-21.8]$ & $3.7\times10^{-11}$\\
\bottomrule
\end{tabular}
\end{table}

Reflection achieves 56.0\% recovery in the earlier core evaluation, compared with 70.6\% for Resample, and uses more completion tokens and wall-clock time. In the subsequent evaluation using the original generation instruction, recovery is 25.0\% for Reflection Repair and 21.4\% for Direct+Compiled, compared with 89.7\% for Compiled. Editing produces smaller program changes than fresh generation. Because the editing comparisons change code exposure and instruction together, the cause of the lower recovery remains unresolved.

\subsection{Static guidance from the first attempt}
\label{app:static-start}

The Static-from-Start study compares overall success across 700 shared fresh scenes, using six executions per condition, neutral retry instructions, and fixed template text (Table~\ref{tab:e2e}). Resample, Compiled, and Fixed Task Template share attempt~0 and its 565 failures, whereas Static-from-Start has 442 initial failures. Conditional recovery is therefore reported descriptively.

The standard initial prompt ends with ``Write a Python program using ONLY the API above.'' Static-from-Start instead uses the complete retry prompt, including the longer neutral instruction in Appendix~\ref{app:prompt-format} and the historical ``attempt 0 failed'' header. It changes initial guidance and the generation instruction together. Their individual contributions remain unresolved.

\begin{table}[!htbp]
\centering
\caption{\textbf{Static-from-Start increases initial success and reduces tokens per solved episode.} Results on 700 fresh scenes with neutral retry instructions. Success and recovery are percentages; recovery uses different initial-failure subsets.}
\label{tab:e2e}
\centering
\small

\begin{tabular}{@{}lrrrr@{}}
\toprule
Condition & Initial & Overall@6 & Recovery@5 & Tok/solved\\
\midrule
Static-from-Start & 36.9 & 92.4 & 88.0 & 2{,}087\\
Fixed Task Template & 19.3 & 90.1 & 87.8 & 2{,}564\\
Compiled & 19.3 & 89.6 & 87.1 & 2{,}578\\
Resample & 19.3 & 73.1 & 66.7 & 3{,}623\\
\bottomrule
\end{tabular}
\end{table}

Static-from-Start achieves $+2.9$\,pp higher overall success than Compiled (95\% CI $[+0.0,+5.7]$, Holm-adjusted $p{=}0.13$). Its advantage over Fixed Task Template is $+2.3$\,pp ($[-0.4,+5.0]$, unadjusted $p{=}0.13$). An exploratory replay under fixed token budgets gives an advantage over Compiled of $+17.6$\,pp at 2{,}000 tokens ($[+13.0,+22.1]$) and $+4.6$\,pp at 5{,}000 ($[+1.1,+8.0]$).

The earlier end-to-end study uses 700 separate scenes and the original generation instruction. Attempt~0 succeeds on 147 scenes, leaving 553 shared failures. Compiled reaches 92.9\% overall success, compared with 73.1\% for Resample and 95.6\% for Fixed Task Template. The template comparison is exploratory.

\clearpage
\subsection{Generalization and failure cases}
\label{sec:app-generalization}
\label{app:breadth}
\label{app:32b-scale}

Across the seven eligible Qwen2.5-Coder-3B/7B settings, Compiled improves recovery over Resample by $+18.1$\,pp (95\% CI $[+13.3,+22.9]$). Its difference from Other-Episode Feedback is smaller, at $+2.3$\,pp ($[-1.1,+5.6]$). The additional studies below are excluded from this combined estimate.

Recovery gains vary across API and perturbation settings. With the position-only API, Compiled improves recovery over Resample by $+30.0$\,pp on cube-to-zone. Both methods reach 93.3\% recovery on place-in-bin. Across three generation seeds, Compiled exceeds Resample in all six tested seed--setting pairs ($+10.7$ to $+26.7$\,pp). With 5\,mm pose noise, the difference is $+16.7$\,pp on lift and $+10.0$\,pp on stack, with only the lift interval excluding zero. At temperature 0.2 on lift with pose shift, Compiled's recovery is $19.6$\,pp lower than Resample's (90\% CI $[-28.6,-10.7]$).

\begin{figure}[!htbp]
\centering
\includegraphics[width=0.70\linewidth]{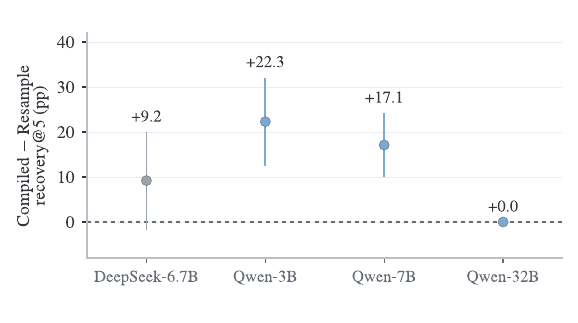}
\caption{\textbf{Recovery differences between Compiled and Resample across model-specific studies.} Each point compares Compiled with Resample within a separately evaluated model and task set.}
\label{fig:app-model-regimes}
\end{figure}

The model-specific results cover different levels of initial success (Figure~\ref{fig:app-model-regimes}). DeepSeek-Coder-6.7B falls below the minimum initial-success rate required for inclusion on both tasks. Its combined recovery is 50.8\% for Compiled and 41.7\% for Resample ($n{=}120$; difference $+9.2$\,pp, $[-1.7,+20.0]$). The separate door-opening study has 1.3\% initial success and also falls below the minimum required for inclusion.

With Qwen2.5-Coder-32B, all 80 lift scenes succeed initially. On stack, 105 of 160 initial attempts succeed, and Compiled and Resample solve the same 52 of the remaining 55 failures (94.5\%). The paired interval has zero width because the methods produce identical outcomes on these 55 failures. Equivalence on other evaluation sets remains untested.

On nut insertion, Compiled recovers 82.5\% of initial failures, compared with 97.5\% for Resample ($n{=}40$; $-15.0$\,pp, 95\% CI $[-27.5,-2.5]$). The decrease is consistent with the center-grasp instruction conflicting with the required rim grasp. Other-Episode Feedback reaches 62.5\% recovery. Attempt validity remains similar for Compiled and Resample (85.1\% and 84.9\%).

\clearpage
\section{Language Feedback Templates}
\label{app:full-library}
The following schema-3.0 texts are extracted directly from \path{src/roborepair/plan/constraints.py}. All 14 label-specific templates, the generic fallback, and both API-variant substitution blocks are included.
\begin{promptpart}{syntax\_error}{cBgPlan}
Diagnosis: The previous program did not compile.
MUST: write a complete syntactically valid program
NEVER: truncated statements
NEVER: markdown or prose inside the code
\end{promptpart}

\begin{promptpart}{timeout}{cBgPlan}
Diagnosis: The previous program exceeded the execution time budget.
MUST: remove unbounded loops and retry/polling logic; write a short direct sequence of API calls
NEVER: while-loops that wait or retry
NEVER: sleep or polling
\end{promptpart}

\begin{promptpart}{invalid\_api}{cBgPlan}
Diagnosis: The previous program called functions that do not exist.
MUST: use ONLY the documented API functions, already defined (no imports, no redefinitions)
NEVER: import statements
NEVER: calling functions not in the API doc
NEVER: redefining API functions
\end{promptpart}

\begin{promptpart}{no\_api\_calls}{cBgPlan}
Diagnosis: The previous program executed but made no API calls.
MUST: make the API calls execute at top level; call any helper function you define
NEVER: defining functions without calling them
\end{promptpart}

\begin{promptpart}{target\_not\_localized}{cBgPlan}
Diagnosis: The previous program never located the target object.
MUST: look up each object with get_object_pose(name) and move using the returned position
NEVER: referencing objects not in the scene
NEVER: guessing positions
\end{promptpart}

\begin{promptpart}{missing\_grasp}{cBgPlan}
Diagnosis: The previous program never attempted a grasp.
MUST: grasp the object before lifting/placing: goto_pose to the object's centre position, then close_gripper()
NEVER: lifting or placing before a successful grasp
\end{promptpart}

\begin{promptpart}{failed\_grasp}{cBgPlan}
Diagnosis: The grasp was attempted but did not secure the object.
MUST: grasp at the object's CENTRE: goto_pose to the object's centre position, then close_gripper(), with the gripper empty
NEVER: closing the gripper away from the object's centre
\end{promptpart}

\begin{promptpart}{grasp\_without\_goto}{cBgPlan}
Diagnosis: The gripper closed while positioned away from the object.
MUST: move the gripper to the object's centre BEFORE closing: goto_pose to the object's centre position, then close_gripper()
NEVER: close_gripper without first moving to the object
\end{promptpart}

\begin{promptpart}{grasp\_while\_holding}{cBgPlan}
Diagnosis: A grasp failed because the gripper was still holding another object.
MUST: grasp only ONE object; move the held object to its destination instead of grasping another
NEVER: closing the gripper while already holding an object
\end{promptpart}

\begin{promptpart}{object\_not\_lifted}{cBgPlan}
Diagnosis: The object was grasped but never lifted.
MUST: after the grasp, LIFT the object: {lift_hint}
NEVER: ending the program with the object still on the table
\end{promptpart}

\begin{promptpart}{released\_after\_lift}{cBgPlan}
Diagnosis: The object was lifted but then released, so it was no longer held at the end.
MUST: do NOT open the gripper after lifting; keep holding (or place on the target first): {place_hint}
NEVER: open_gripper immediately after lifting
\end{promptpart}

\begin{promptpart}{bad\_placement}{cBgPlan}
Diagnosis: The object was lifted but not placed correctly on the target.
MUST: place the object ON the target: {place_hint}
NEVER: releasing the object away from the target
\end{promptpart}

\begin{promptpart}{unsafe\_path}{cBgPlan}
Diagnosis: An object was knocked off the table.
MUST: approach objects from above with z_approach; avoid sweeping motions near objects
NEVER: motions that knock objects off the table
\end{promptpart}

\begin{promptpart}{door\_not\_opened}{cBgPlan}
Diagnosis: The handle was grasped but the door was never pulled open.
MUST: after gripping the handle, PULL it sideways away from the door frame: pull('door_handle', dx, dy) with a horizontal vector of at least 0.15 m
NEVER: releasing the handle without pulling it
NEVER: lifting the handle instead of dragging it sideways
\end{promptpart}

\begin{promptpart}{unknown\_failure (fallback)}{cBgPlan}
Diagnosis: The task did not succeed.
MUST: fix the cause of failure and complete every task step
\end{promptpart}

\begin{promptpart}{API tier M substitutions}{cBgApi}
lift_hint: call lift(name, height=0.1) after the grasp
place_hint: use place_on(source, target) as the final action (it releases)
\end{promptpart}

\begin{promptpart}{API tier P substitutions}{cBgApi}
lift_hint: after closing the gripper, goto_pose to a position 0.15 m ABOVE the grasp point with z_approach=0 (the object moves with the gripper)
place_hint: while holding, goto_pose above the target position, then open_gripper()
\end{promptpart}

\end{document}